\documentclass[conference]{IEEEtran}
\IEEEoverridecommandlockouts
\usepackage{cite}
\usepackage{amsmath,amssymb,amsfonts}
\usepackage{algorithmic}
\usepackage{graphicx}
\usepackage{textcomp}
\usepackage{xcolor}
\usepackage{booktabs}
\usepackage{multirow}
\usepackage[T1]{fontenc}
\usepackage{times}
\usepackage[colorlinks=true,
            linkcolor=blue,
            urlcolor=blue,
            citecolor=blue]{hyperref}

\usepackage{newtxmath} 

\def\BibTeX{{\rm B\kern-.05em{\sc i\kern-.025em b}\kern-.08em
    T\kern-.1667em\lower.7ex\hbox{E}\kern-.125emX}}
\begin{document}

\title{Biomechanical-phase based Temporal Segmentation in Sports Videos: a Demonstration on Javelin-Throw\\
\thanks{Supported by Ministry of Youth Affairs and Sports having Project No. K-15015/16/2023-SP.V}
}

\author{\IEEEauthorblockN{1\textsuperscript{st} Bikash Kumar Badatya}
\IEEEauthorblockA{\textit{Department of Physics} \\
\textit{Indian Institute of Technology Gandhinagar}\\
Gandhinagar, India \\
bikash.badatya@iitgn.ac.in}
\and
\IEEEauthorblockN{2\textsuperscript{nd} Vipul Baghel}
\IEEEauthorblockA{\textit{Department of Electrical Engineering} \\
\textit{Indian Institute of Technology Gandhinagar}\\
Gandhinagar, India \\
baghelvipul@iitgn.ac.in}
\and
\IEEEauthorblockN{3\textsuperscript{rd} Jyotirmoy Amin}
\IEEEauthorblockA{\textit{Department of Mathematics} \\
\textit{IISER Berhampur}\\
Brahmapur, India \\
jyotirmoy22@iiserbpr.ac.in}
\and
\hspace{1em}\IEEEauthorblockN{4\textsuperscript{th} Ravi Hegde}
\hspace{50em}\IEEEauthorblockA{
\textit{Department of Electrical Engineering} \\
\textit{Indian Institute of Technology Gandhinagar} \\
Gandhinagar, India \\
hegder@iitgn.ac.in}
}

\maketitle

\begin{abstract}
Precise analysis of athletic motion is central to sports analytics, particularly in disciplines where nuanced biomechanical phases directly impact performance outcomes. Traditional analytics techniques rely on manual annotation or laboratory-based instrumentation -- which are time-consuming, costly, and lack scalability. Automatic extraction of relevant kinetic variables requires a robust and contextually-appropriate temporal segmentation. Considering the specific case of elite javelin-throw, we present a novel unsupervised framework for such a contextually-aware segmentation which applies the structured optimal transport (SOT) concept to augment the well-known Attention-based Spatio-Temporal Graph Convolutional Network (ASTGCN). This enables the identification of motion phase transitions without requiring expensive manual labeling. Extensive experiments demonstrate that our approach outperforms state-of-the-art unsupervised methods, achieving 71.02\% mean average precision (mAP) and 74.61\% F1-score on test data—substantially higher than competing baselines. We also release a new dataset of 211 manually-annotated professional javelin-throw videos with frame-level annotations, covering key biomechanical phases: approach steps, drive, throw, and recovery. 
\end{abstract}



\section{Introduction}

Understanding the temporal structure of athletic motion is vital for performance assessment, skill refinement, and coaching feedback. In high-performance sports like javelin throw, even subtle biomechanical differences can significantly affect outcomes such as velocity, distance, and accuracy. \textit{Fine-grained action segmentation}—identifying precise temporal boundaries between motion phases—enables such complex movements to be decomposed into interpretable components~\cite{piergiovanni2017learning}.

In particular, throwing disciplines (e.g., javelin, discus, shot put) follow sevral sequential biomechanical phases: \textit{approach}, \textit{transition}, \textit{delivery}, \textit{release}, and \textit{recovery}~\cite{trasolini2022biomechanical}. For instance, in javelin, these stages involve a high-speed run-up, a blocking step that redirects momentum, and a release followed by recovery. Each phase uniquely contributes to energy transfer, alignment, and throw optimization. However, inter-athlete variability in stride length, joint timing, and run-up speed—driven by physique and technique—necessitates adaptive, individualized motion analysis frameworks~\cite{trasolini2022biomechanical,best1993three}. Conventional performance analysis in throwing sports often depends on manual annotation or laboratory-based instrumentation, such as high-speed cameras and motion capture systems. While these approaches can yield detailed biomechanical insights, they are inherently time-consuming, expensive, and unsuitable for deployment at scale. More critically, they fail to generalize across varying environments and athlete-specific techniques. Automatic extraction of meaningful kinetic variables—such as release angle, momentum buildup, and joint sequencing—thus requires not only accurate data but also a robust and context-aware method for \textit{temporal action segmentation}. This motivates the development of automated, adaptive frameworks capable of parsing motion into semantically rich phases, without reliance on cumbersome or expert-driven annotation pipelines.

\textbf{Temporal action localization (TAL)}~\cite{shao2023action}—identifying meaningful segments within continuous motion—enables downstream tasks such as \textit{style classification}, \textit{technique correction}, and \textit{performance tracking}~\cite{chen2020action,girdhar2019video}. Isolated segments allow for focused visualization, cross-session alignment, and targeted feedback, while reducing data redundancy. Though, most existing analyses rely on labor-intensive manual annotations or lab setups, limiting scalability. Prior studies using 3D motion capture~\cite{liu2010sequences,kohler2020biomechanical} offer biomechanical insights but are constrained by small cohorts and lack automated temporal segmentation from pose or video data.
\footnote{This paper has been accepted at the IEEE STAR Workshop 2025.}
Fully supervised methods~\cite{farha2019ms,lu2024fact} perform well but require dense frame-level labels, which are expensive and impractical for sports with high variability. In contrast, unsupervised approaches~\cite{kukleva2019unsupervised,xu2024temporally,kumar2022unsupervised} bypass this dependency and offer scalable alternatives for segmenting complex, variable motion like javelin throw. By detecting motion transitions in untrimmed videos without supervision, such systems enable real-time feedback, long-term performance tracking, and accessibility in both elite and grassroots settings. Despite their promise, unsupervised segmentation methods remain underexplored in throwing sports~\cite{xu2024temporally}.

TAL can be performed on two primary formats of video data: \textit{RGB-based} and \textit{skeleton-based}. RGB-based TAL methods operate directly on raw visual inputs such as frames, optical flow, or pre-extracted visual features. These methods benefit from rich appearance cues and contextual information, which are essential for recognizing subtle scene dynamics. However, they are often sensitive to background clutter, lighting variations, occlusions, and inter-class visual ambiguity. In contrast, skeleton-based TAL leverages human pose representations—typically 2D or 3D joint trajectories—to model the dynamics of human motion in a more structured and interpretable form. 

Skeleton-based TAL frameworks—built on skeletal joint trajectories—offer modality-invariant, interpretable representations robust to appearance changes. These systems support automated detection of key motion events (e.g., release initiation), facilitating objective diagnostics and scalable sports analytics. In summary, our contributions are:
\begin{itemize}
    \item \textbf{Unsupervised Skeleton-Based Segmentation:} We propose a novel framework combining a denoising-based ASTGCN encoder with a structured optimal transport module to segment javelin throw phases using 2D pose data. Our model is lightweight, interpretable, and achieves competitive performance.
    \item \textbf{Javelin Pose Dataset:} We introduce the first frame-level annotated dataset for javelin throw segmentation, extracted from real-world competition videos, capturing diverse styles and conditions. The dataset is available on \url{https://github.com/Bikudebug/Javelin_Throw_Dataset}.
\end{itemize}

\section{Related Work}

\subsection{Model-Based Biomechanical Analysis}
Early sports motion studies used inverse dynamics and biomechanical models based on motion capture and joint kinetics~\cite{liu2010sequences,kohler2020biomechanical}. While informative for joint sequencing and energy transfer, these methods depend on expensive lab setups, small cohorts, and manual annotations, limiting real-world scalability. This motivates learning-based approaches that generalize beyond controlled environments.

\subsection{Fully Supervised Temporal Action Segmentation}
Supervised methods like TCN~\cite{lea2017temporal} and MS-TCN~\cite{farha2019ms} model long-range dependencies using dense labels. More recent transformer-based models~\cite{neimark2021video,lu2024fact} further enhance context modeling. However, their reliance on large-scale frame-level annotations makes them impractical for sports like javelin, where labeled data is scarce and performance varies widely. This underscores the need for label-free alternatives.

\subsection{Unsupervised Action Segmentation}
Unsupervised methods bypass manual labels using proxy objectives like timestamp prediction~\cite{kukleva2019unsupervised}, contrastive learning~\cite{zhuang2020unsupervised,swetha2021unsupervised}, and probabilistic alignment~\cite{li2021action,ding2023temporal}. Optimal transport-based methods such as TOT~\cite{kumar2022unsupervised} and UFSA~\cite{tran2024permutation} offer structured alignment but assume fixed class balance or action order, which limits applicability in diverse, athlete-specific throwing motions. We overcome this by applying structured optimal transport without enforcing rigid sequence constraints.

\subsection{Pose-Based and Skeleton-Centric Action Segmentation}
Skeleton-based approaches are robust to appearance changes and occlusions, making them suitable for scalable action segmentation. Prior works use contrastive learning~\cite{tian2025stitch}, latent temporal modeling~\cite{yang2023lac}, or multi-modal skeleton heatmaps~\cite{hyder2024action} to capture pose dynamics. These advances demonstrate the interpretability and generalizability of pose-based representations. Our method builds on this by combining spatio-temporal graph embeddings with optimal transport for unsupervised segmentation of throwing phases.

\begin{figure*}[!t]
  \centering
  \includegraphics[width=\textwidth,keepaspectratio]{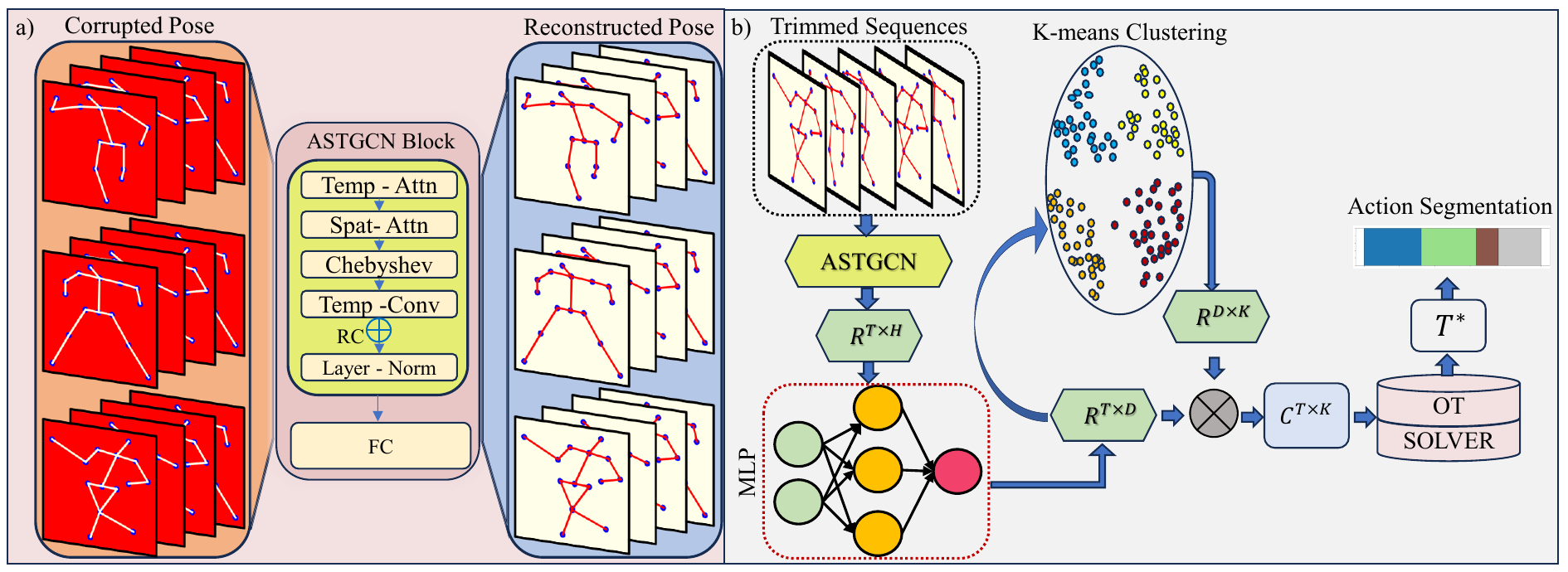}
  \caption{Overview of the proposed framework.
\textbf{(a)} A spatio-temporal encoder based on ASTGCN~\cite{guo2019attention}  reconstructs clean 2D pose sequences from corrupted inputs using Chebyshev graph convolutions, temporal/spatial attention, residual connections, and a fully connected (FC) projection layer. 
\textbf{(b)} The high-dimensional representations \( R^{T \times H} \) are first reduced via a lightweight MLP to \( R^{T \times D} \), then clustered using K-means. A visual cost matrix \( C^{T \times K} \) is computed as input to the Optimal Transport (OT) solver, which generates a soft alignment matrix \( T^* \) used as pseudo-labels for action segmentation.}
  \label{fig:arch}
\end{figure*}

\section{Methodology}

We propose an unsupervised framework for segmenting fine-grained motion
phases in untrimmed javelin throw videos using 2D pose sequences. The proposed method consists of two stages: first, a graph model is trained through a denoising task that reconstructs clean joint trajectories from artificially corrupted inputs; second, the resulting spatio-temporal embeddings are post-processed using a structured optimal transport formulation to detect motion phase transitions without any manual supervision~\cite{xu2024temporally}. As illustrated in Fig.~\ref{fig:arch}.

We employ ASTGCN~\cite{guo2019attention} to encode 2D pose sequences into motion-sensitive spatio-temporal embeddings. Each input sample \( P \in \mathbb{R}^{B \times W \times J \times 2} \) comprises a batch of pose sequences with \( B \) as batch size, \( W \) the temporal window length, \( J \) the number of joints, and 2D coordinates. During training, Gaussian noise is added to enhance robustness. Each ASTGCN block integrates joint-wise spatial attention, Chebyshev spectral graph filtering, and temporal convolution, followed by residual connections and layer normalization for stability. The final block outputs latent motion representations, which are projected back to the original pose space via a fully connected (FC) layer to reconstruct denoised coordinates (see Fig.~\ref{fig:arch}(a)).

The network is optimized using a combination of reconstruction and velocity consistency losses (Eq.\eqref{eq:totalloss}):
\begin{gather}
\mathcal{L}_{\text{MSE}} = \frac{1}{BWJ} \sum_{b=1}^{B} \sum_{t=1}^{W} \sum_{j=1}^{J} \left\| \hat{P}_{b,t,j} - P_{b,t,j} \right\|_2^2, \tag{1} \label{eq:mse} \\[6pt]
v^{\text{pred}}_{b,t,j} = \hat{P}_{b,t,j} - \hat{P}_{b,t-1,j}, \quad
v^{\text{gt}}_{b,t,j} = P_{b,t,j} - P_{b,t-1,j} \notag \\[4pt]
\mathcal{L}_{\text{vel}} = \frac{1}{BWJ} \sum_{b=1}^{B} \sum_{t=2}^{W} \sum_{j=1}^{J} \left\| v^{\text{pred}}_{b,t,j} - v^{\text{gt}}_{b,t,j} \right\|_2^2, \tag{2} \label{eq:velloss} \\[6pt]
\mathcal{L}_{\text{total}} = \mathcal{L}_{\text{MSE}} + \lambda_{\text{vel}} \cdot \mathcal{L}_{\text{vel}}. \tag{3} \label{eq:totalloss}
\end{gather}

Here, \( \lambda_{\text{vel}} \) balances spatial reconstruction and temporal consistency during training.

\subsection{Feature Extraction}

At inference time, the trained ASTGCN model processes clean pose sequences to extract high-dimensional spatio-temporal embeddings. These outputs, denoted as \( X \in \mathbb{R}^{T \times H} \), capture localized motion patterns and joint interactions across frames, where \( H \) is the hidden feature dimension. To obtain compact and semantically meaningful embeddings suitable for segmentation, the high-dimensional features are passed through a lightweight multi-layer perceptron (MLP) with one hidden layer. This projection yields the final embedding sequence \( X \in \mathbb{R}^{T \times D} \), where \( D \) is the dimensionality of the latent representation space. These MLP-projected embeddings serve as input to the structured optimal transport~\cite{thorpe2021intro} module, as shown in Fig.~\ref{fig:arch}(b).

\subsection{Unsupervised Segmentation via Structured Optimal Transport}

To decode temporal segments and identify motion transitions from the embedding sequence \( X \in \mathbb{R}^{T \times D} \), we formulate the problem within the framework of optimal transport on structured data~\cite{kumar2022unsupervised}. This allows us to model segmentation as a soft alignment between frame-level embeddings and latent motion prototypes, while preserving temporal consistency.

Let \( A = \{a_1, \dots, a_K\} \subset \mathbb{R}^D \) denote a set of learnable prototypes representing \( K \) latent motion classes. A visual cost matrix \( C^{\text{vis}} \in \mathbb{R}^{T \times K} \) is computed using cosine distances:
\begin{equation}
C^{\text{vis}}_{t,k} = 1 - \frac{x_t^\top a_k}{\|x_t\| \|a_k\|}.
\tag{4}
\label{cost}
\end{equation}

To regularize temporal coherence, we define a temporal structure matrix \( C^{\text{temp}} \in \mathbb{R}^{T \times T} \) and a category-wise matrix \( C^{\text{cat}} \in \mathbb{R}^{K \times K} \) that penalize abrupt label changes. The segmentation task is formulated as an unbalanced optimal transport~\cite{chizat2016scaling,sejourne2021unbalanced,thual2022aligning} problem over the coupling matrix \( T \in \mathbb{R}^{T \times K} \), optimized using:
\begin{equation}
\begin{split}
\mathcal{F}(T) &= \alpha \cdot \langle C^{\text{temp}} T C^{\text{cat}}, T \rangle + (1 - \alpha) \cdot \langle C^{\text{vis}}, T \rangle \\
&\quad + \lambda \cdot \mathrm{KL}(T^\top \mathbf{1}_T \,\|\, q),
\end{split}
\tag{5}
\label{eq:frobenius}
\end{equation}

where \( \langle A, B \rangle := \sum_{i,j} A_{i,j} B_{i,j} \) denotes the inner product of Frobenius. The first term in Eq.\eqref{eq:frobenius} corresponds to the Fused Gromov-Wasserstein optimal transport~\cite{li2023efficient,thual2022aligning,titouan2019optimal,vayer2020fused}, aligning the temporal structure across frames. The second term in Eq.\eqref{eq:frobenius} is the Kantorovich OT~\cite{thorpe2021intro} that enforces the consistency of the appearance level. The third term in Eq.\eqref{eq:frobenius} is a Kullback--Leibler divergence:
\begin{equation}
\text{KL}(T^\top \mathbf{1}_T \,\|\, q) = \sum_{k=1}^{K} \left( T^\top \mathbf{1}_T \right)_k \log \left( \frac{(T^\top \mathbf{1}_T)_k}{q_k} \right),
\tag{6}
\label{eq:kl}
\end{equation}
where \( q \in \mathbb{R}^K \) is the prior class distribution. Hyperparameters \( \alpha \in [0, 1] \) and \( \lambda \) balance structure, appearance, and marginal relaxation. To solve this, we use a mirror descent-based solver with entropy regularization. The optimal coupling matrix \( T^* \in \mathbb{R}^{T \times K} \) yields soft frame-to-class assignments.


Final frame-level labels are obtained by selecting the class with the highest assignment probability:
\begin{equation}
\hat{y}_t = \arg\max_k T^*_{t,k}.
\tag{7}
\label{eq:prob}
\end{equation}
This produces temporally coherent segmentations that reflect the latent phase structure of complex actions in an unsupervised manner.

\subsection{Unsupervised Learning via Pseudo-Labeling}

The optimal transport plan also enables self-supervised learning~\cite{kukleva2019unsupervised}. For each sequence \( b \) in a mini-batch, a pseudo-label matrix \( P^b \in \mathbb{R}^{T \times K} \) is constructed by normalizing rows of the transport matrix:
\begin{equation}
P^b_{i,j} := \frac{T^{*b}_{i,j}}{\sum_{j'=1}^{K} T^{*b}_{i,j'}},
\tag{8}
\label{eq:pseudo}
\end{equation}
ensuring \( P^b \in \Delta_T^K \), i.e., each row is a valid probability distribution over motion classes.

These soft pseudo-labels are used to supervise the embedding network via an entropy-based loss. Let \( f_\theta(x_i^b)_j \) denote the softmax prediction for class \( j \), then:
\begin{equation}
\mathcal{L}_{\text{train}}(\theta) = - \sum_{b=1}^{B} \sum_{i=1}^{T} \sum_{j=1}^{K} P^b_{i,j} \log f_\theta(x_i^b)_j.
\tag{9}
\label{eq:ltrain}
\end{equation}

This establishes a closed unsupervised learning loop: improved embeddings yield better alignment via optimal transport, which in turn provides refined supervision for representation learning—all without any ground-truth labels.

\section{Experimental Setup}

\subsection{Dataset}

In this work, we address the lack of javelin-throw datasets by compiling a collection of 211 videos, 111 men’s and 100 women’s sourced from YouTube and official Olympic recordings of major events including the Paris Olympics 2024, the European Athletics Championships (Rome 2024), the World Athletics Championships (Oregon 2022), the Tokyo Olympics 2020, and the IAAF World Championships 2015. We employed the PyTorch-based MMPose~\cite{sengupta2020mm} framework for pose detection, extracting 16‑joint skeletons in MPII format. We manually annotated key frames to serve as the ground-truth segmentation boundaries. The data set is labeled at the frame level in four core action phases: steps, drive, throw, and recovery. Here, the 'steps phase' covers the rhythmic approach strides, the 'drive phase' begins after the last crossover step and involves an aggressive forward motion to generate momentum for the throw, the 'throw phase' captures the javelin release, and the 'recovery phase' consists of post-throw deceleration and body stabilization. This annotation provides a reliable basis for training and evaluating unsupervised segmentation models in fine-grained sports motion analysis.

\subsection{Evaluation Metrics}

To assess the quality of unsupervised segmentation~\cite{kukleva2019unsupervised,kumar2022unsupervised,li2021action,swetha2021unsupervised,vidalmata2021joint,tran2023permutation}, we evaluate our method using frame-level ground truth labels by aligning the predicted cluster assignments with true action labels through Hungarian matching. In our setup, the matching is performed \textit{once across the entire dataset}, enforcing consistent cluster-to-label mappings across all videos. We report four commonly used metrics: mean over frames (MoF), F1-score, mean intersection-over-union (mIoU), and mean average precision (mAP) ~\cite{swetha2021unsupervised,vidalmata2021joint}.

\subsection{Implementation Details}

The ASTGCN~\cite{guo2019attention} encoder comprises 3 stacked blocks, each incorporating joint-wise spatial attention, a Chebyshev graph convolution with 7 filters, and a temporal convolution to capture frame-wise dynamics. To encourage robust learning, Gaussian noise with a random factor of 0.1 is added to the 2D pose inputs during training. Feature representations extracted from the final ASTGCN block are high-dimensional vectors of size 64. These embeddings are passed through a lightweight MLP encoder with a single hidden layer, which projects them into a lower-dimensional latent space of size 40 suitable for structured optimal transport.

The ASTGCN is trained using the Adam optimizer with a learning rate of \(1 \times 10^{-4}\). The encoder MLP is optimized separately with the Adam optimizer using a learning rate of \(1 \times 10^{-3}\) and a weight decay of \(1 \times 10^{-4}\). Latent motion prototypes are initialized via k-means clustering on a randomly selected subset of frame embeddings. During training, each video is divided into 5 uniformly spaced temporal bins, and one frame is sampled at random from each bin to ensure temporal diversity while reducing computational overhead. The number of motion segments \(K\) is set to the ground truth number of phases per activity category, consistent with prior work~\cite{kukleva2019unsupervised,kumar2022unsupervised,li2021action}.

\begin{figure*}[!t]
  \centering
  \includegraphics[width=\textwidth,keepaspectratio]{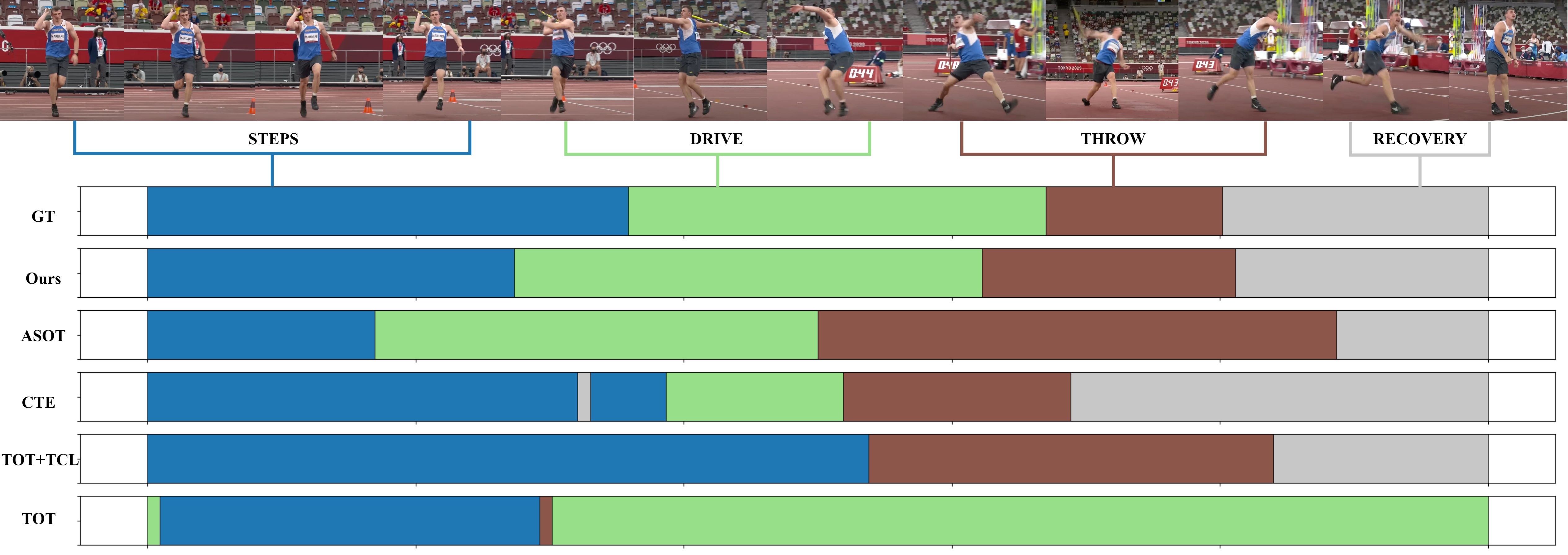}
  \caption{Qualitative comparison of action segmentation results on a javelin throw sequence. The top row shows selected video frames with annotated ground truth phases: Steps, Drive, Throw, and Recovery. The bar plots below visualize the phase segmentation results from different methods. Compared to baseline approaches—TOT~\cite{kumar2022unsupervised}, TOT+TCL~\cite{kumar2022unsupervised}, CTE~\cite{kukleva2019unsupervised}, and ASOT~\cite{xu2024temporally}—our method demonstrates superior phase alignment, accurately detecting transitions and segment lengths that closely match the ground truth.}
  \label{fig:segment}
\end{figure*}
\begin{table}[ht]
  \centering
  \setlength{\tabcolsep}{4pt} 
  \renewcommand{\arraystretch}{1.2} 
  \caption{Comparison with state-of-the-art methods under the unsupervised (full) evaluation protocol. We report mAP, F1-score, mIoU, and MoF on training and test sets. Our method consistently outperforms baselines across all metrics.}
  \label{tab:seg_performance}
  \footnotesize
  \resizebox{\columnwidth}{!}{
    \begin{tabular}{lcccccccc}
      \toprule
      \multirow{2}{*}{Method} 
       & \multicolumn{4}{c}{Train} 
       & \multicolumn{4}{c}{Test} \\
      \cmidrule(lr){2-5}\cmidrule(lr){6-9}
        & mAP & F1 & mIoU & MoF 
        & mAP & F1 & mIoU & MoF \\
      \midrule
      TOT~\cite{kumar2022unsupervised}      & 34.00 & 34.91 & 22.20 & 35.83 & 33.32 & 33.70 & 20.21 & 34.90 \\
      TOT+TCL~\cite{kumar2022unsupervised}  & 37.50 & 47.95 & 40.18 & 33.63 & 39.30 & 50.03 & 41.32 & 50.87 \\
      CTE~\cite{kukleva2019unsupervised}      & 54.57 & 66.85 & 48.84 & 65.20 & 54.82 & 62.74 & \textbf{50.58} & 63.42 \\
      ASOT~\cite{xu2024temporally}     & 61.00 & \textbf{70.16} & 45.46 & 62.50 & 60.20 & 57.55 & 41.80 & 61.19 \\
      \textbf{OURS} & \textbf{76.77} & 56.60 & \textbf{52.01} & \textbf{69.70} & \textbf{71.02} & \textbf{74.61} & 48.01 & \textbf{64.04} \\
      \bottomrule
    \end{tabular}%
  }
\end{table}
\section{Results}

\subsection{State-of-the-Art Comparison}

We report a comparative evaluation of our method against recent state-of-the-art approaches in Table~\ref{tab:seg_performance}. All results are reported under the \textit{unsupervised full} evaluation protocol, where model predictions are aligned to ground truth labels globally across the entire dataset using Hungarian matching. We compare our approach with a range of recent state-of-the-art baselines, including transformer-based and graph neural network-based segmentation frameworks.

TOT~\cite{kumar2022unsupervised} approaches the task by jointly learning frame representations and clustering them in an online manner, relying on memory bank sampling and cluster assignment regularization. TOT+TCL~\cite{kumar2022unsupervised} builds upon this by adding a temporal coherence loss (TCL), enforcing smoother transitions across frame clusters. CTE~\cite{kukleva2019unsupervised} introduces continuous temporal embeddings learned via contrastive objectives, where segment similarity is encouraged using temporal proximity. ASOT~\cite{xu2024temporally} applies a structured optimal transport formulation over pose-based embeddings, combining visual and structural alignment objectives.

As shown in Table~\ref{tab:seg_performance}, TOT and TOT+TCL perform poorly across metrics. While CTE performs relatively well on the mIoU metric for test data (50.58\%), it lags in overall scores. ASOT demonstrates strong performance on training data, particularly in terms of F1-score (70.16\%), but its effectiveness drops noticeably on test data, indicating limited generalization capability. In contrast, our method achieves the highest scores across the board, including 76.77\% mAP on training and 71.02\% on test, significantly outperforming all baselines in both consistency and generalization. These improvements are also visually supported in Fig.~\ref{fig:segment}, where our model produces more accurate and temporally coherent segmentations. Our method consistently achieves the best performance across all four metrics, with relative improvements of 27.06\% in mAP, 12.61\% in F1-score, 11.19\% in mIoU, and 15.93\% in MoF over the next best average scores, highlighting its substantial advantage over existing baselines.

\begin{table}[ht]
\centering
\renewcommand{\arraystretch}{1.2}
\caption{Ablation study on different GCN-based encoder architectures and training configurations. ASTGCN combined with our proposed unsupervised pipeline achieves superior segmentation performance across all evaluation metrics.}

\label{tab:seg_performance_avg}
\footnotesize
\begin{tabular}{lcccc}
\toprule
Model & mAP & F1 & mIoU & MoF \\
\midrule
AGCN~\cite{lee2020background}     & 69.50  & 48.84 & 44.67 & 62.64 \\
TSAGCN~\cite{shi2020weakly}   & 71.22  & 49.28 & 45.71 & 63.33 \\
STGCN~\cite{min2020adversarial}    & 71.39  & 50.05 & 45.75 & 63.34 \\
\textbf{OURS} & \textbf{73.90} & \textbf{65.61} & \textbf{50.01} & \textbf{66.87} \\
\bottomrule
\end{tabular}
\end{table}
\subsection{Ablation Study}
To evaluate the robustness and generalizability of our framework, we conduct an ablation study focusing on three key components: the choice of graph-based encoder, sliding window configuration, and frame sampling strategy. We integrate AGCN~\cite{lee2020background}, TSAGCN~\cite{shi2020weakly}, and ST-GCN~\cite{min2020adversarial} into our unsupervised segmentation pipeline and benchmark them against our ASTGCN-based model. As shown in Table~\ref{tab:seg_performance_avg}, ASTGCN consistently outperforms the other backbones across all evaluation metrics, validating its ability to effectively capture both spatial joint dependencies and temporal motion patterns through attention-enhanced temporal modeling. Additionally, we explore different sliding window sizes \(W \in \{7, 15, 30, 40\}\) to divide pose sequences into overlapping sub-sequences. We observe that using \(W = 30\) offers the best trade-off between fine-grained motion encoding and adequate contextual information. While shuffled windows aid training diversity, serial windowing at inference preserves temporal consistency.

We also examine the effect of varying the number of sampled frames per video using \(n \in \{5, 7, 30, 100, 256\}\), following the bin-based sampling approach of~\cite{kukleva2019unsupervised}. Initially, performance improves with more frames, but excessive sampling introduces redundancy and overfitting. Surprisingly, sampling only 5 frames—each drawn from evenly spaced temporal bins—achieves the most reliable results, balancing efficiency and diversity in representing different action phases. For all ablation experiments, the number of action segments is fixed at \(K = 4\), consistent with our dataset's annotated phase labels. These findings collectively demonstrate that backbone architecture, temporal window size, and sampling density significantly impact unsupervised segmentation performance.

\section{Conclusion}
In this work, we presented a novel unsupervised framework for fine-grained temporal segmentation of javelin throw videos by combining a denoising-based spatio-temporal graph convolutional encoder with structured optimal transport for phase boundary discovery. Our approach operates directly on 2D pose sequences, enabling interpretable, label-free segmentation of key motion phases. While effective, a current limitation is the need to predefine the number of motion segments, assuming consistent phase counts across videos. In future work, we aim to address this by developing adaptive mechanisms for phase count estimation and extending the framework toward micro-level action segmentation for extracting higher-level biomechanical insights such as stride length and step count. We also plan to explore more advanced backbone architectures to enhance segmentation accuracy and generalize our method to other multi-phase athletic activities beyond javelin, including a broader range of throwing sports.
\section*{Acknowledgment}

We acknowledge the Ministry of Youth Affairs and Sports, Department of Sports, Government of India, for providing financial support to our research in Sports Analytics.

\end{document}